%% file: main.tex
\documentclass{article}
\input{setup/imports}

\title{CreativityNeuro: Steering Language Model Weights to \\Improve Divergent Thinking and Reduce Mode Collapse}

\begin{document}

\ifcolmsubmission
\linenumbers
\fi

\input{sections/01_authors}
\input{sections/02_abstract}

\input{sections/03_introduction}

\input{sections/04_background}

\input{sections/05_method}
\input{sections/06_dat}
\input{sections/07_aut_tt}
\input{sections/08_mode_collapse}
\input{sections/09_dt_vs_factual}
\input{sections/10_limitations}

\bibliography{references}
\bibliographystyle{setup/colm2026_conference}

\input{sections/11_appendix}

\end{document}

%% file: setup/imports.tex
\usepackage[preprint]{setup/colm2026_conference}
\usepackage{microtype}
\usepackage{graphicx}
\usepackage{booktabs}
\usepackage{url}
\usepackage{array}
\usepackage{multirow}
\usepackage{hyperref}
\usepackage{amsmath, amsthm, amsfonts, amssymb}

\usepackage{algorithm, algorithmic}
\usepackage{subcaption}
\usepackage{mathtools}
\usepackage[svgnames]{xcolor}
\usepackage{lineno}
\usepackage[breakable]{tcolorbox}
\usepackage{wrapfig}
\usepackage{yfonts}
\usepackage{titletoc}
\usepackage{placeins}
\usepackage{twemojis}

\definecolor{darkblue}{rgb}{0, 0, 0.5}
\definecolor{batlowwarm}{HTML}{E19752}
\definecolor{batlowmid}{HTML}{5C9344}
\definecolor{batlowcool}{HTML}{134961}
\hypersetup{colorlinks=true, citecolor=darkblue, linkcolor=darkblue, urlcolor=darkblue}

\usepackage[capitalize,noabbrev]{cleveref}

\newtcolorbox{promptbox}[1][]{
  colback=black!3,
  colframe=black!30,
  boxrule=0.4pt,
  arc=1.5pt,
  left=8pt, right=8pt, top=6pt, bottom=6pt,
  fontupper=\small\ttfamily,
  before skip=6pt,
  after skip=6pt,
  title={\small\sffamily\bfseries #1},
  coltitle=black,
  colbacktitle=black!8,
}

\newtcolorbox{takeaway}{
  colback=batlowcool!6,
  colframe=batlowcool!40,
  boxrule=0.4pt,
  arc=1.5pt,
  left=8pt, right=8pt, top=6pt, bottom=6pt,
  fontupper=\small,
  before skip=8pt,
  after skip=8pt,
}

\theoremstyle{plain}

\theoremstyle{definition}

\theoremstyle{remark}

%% file: sections/01_authors.tex

\author{Samuel Schapiro\thanks{Corresponding author: \texttt{schapironietzsche@gmail.com}} \\
Univeristy of Illinois, Urbana-Champaign
\AND
Core Francisco Park \\
Center for Brain Science, Harvard University \\
CBS-NTT Program in Physics of Intelligence, Harvard University \\
Prior Computers \\
\And
Felix Sosa \\
Prior Computers  \\
\And
Lav R. Varshney \\
AI Innovation Institute, Stony Brook University
}

\maketitle

%% file: sections/02_abstract.tex
\begin{figure}[h]
    \centering
    \includegraphics[width=\textwidth]{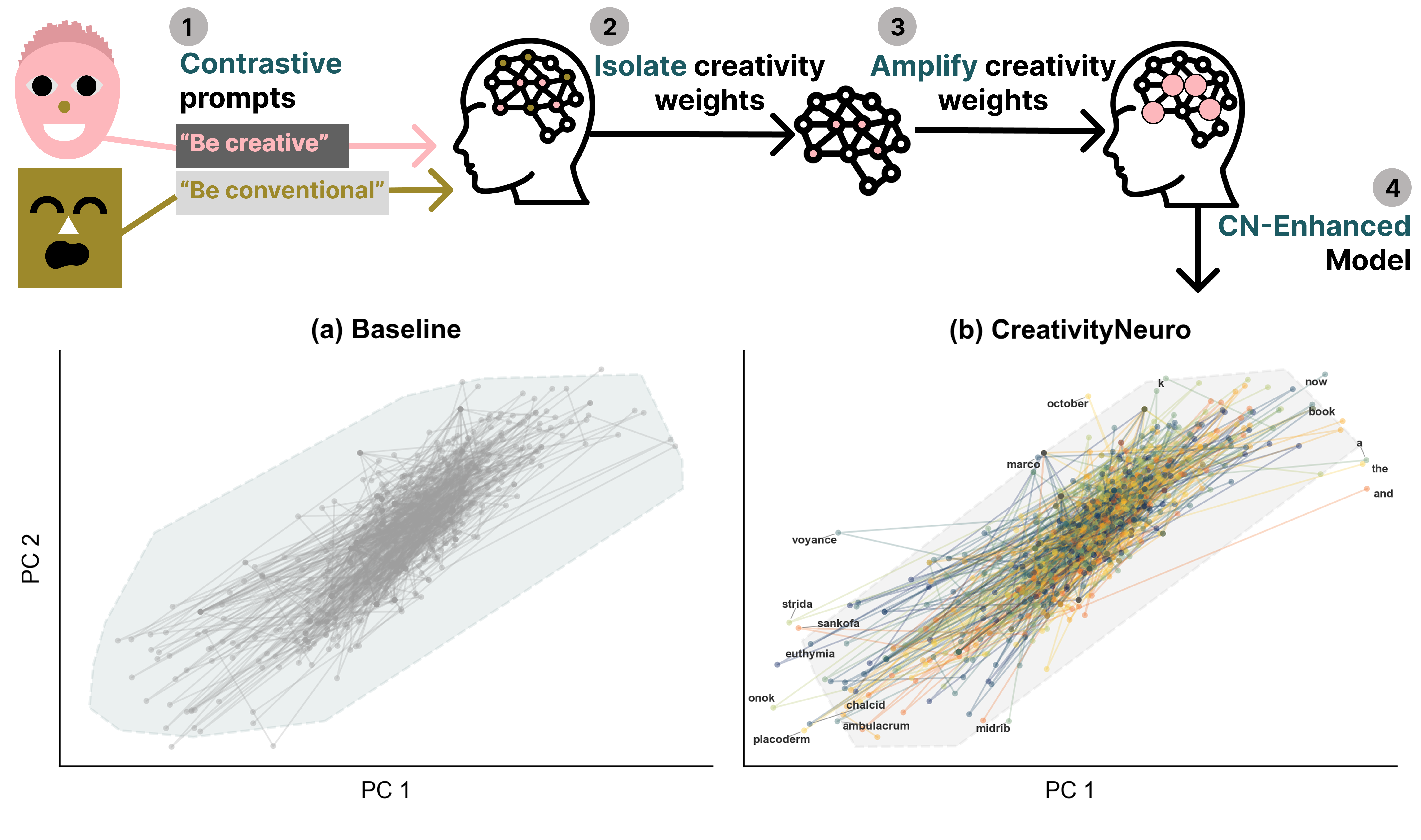}
    \caption{\textbf{CreativityNeuro (CN) pipeline.} Given a pair of contrastive creative prompts, CN computes parameter importance scores, selects a sparse subset of creativity-relevant parameters, and applies a scaled weight perturbation---without requiring behavioral datasets or gradient-based finetuning. CN improves divergent thinking across various tasks. Subplot (b) visualizes CN \emph{thinking outside of the ``box''}(i.e., the convex hull of baseline DAT responses), despite baseline responses falling within CN's convex hull in subplot (a).}
    \label{fig:dat_trajectories}
\end{figure}

\begin{abstract}
Divergent thinking is a crucial aspect of creativity, yet large language models (LLMs) tend to consistently generate similar responses to open-ended questions, in what has been termed the artificial hivemind effect. Here, we introduce CreativityNeuro, a data-free method for enhancing divergent thinking in LLMs via contrastive weight steering. We evaluate our method across multiple creativity assessments and report several main findings. On the Divergent Association Task (DAT), a vocabulary-space creativity test, CreativityNeuro improves performance by up to 14 human percentile points. Next, in a large-scale human evaluation (N=720) on the Alternative Uses Test (AUT) and the Task Task, CreativityNeuro achieves significant improvements in originality, surprise, and creativity, transferring to longer-form and more open-ended tasks. Importantly, we find that across all three tasks, CreativityNeuro demonstrably reduces measures of mode collapse. Moreover, activation steering achieves comparable performance to CreativityNeuro on the DAT, but it does not transfer to the AUT and Task Task, demonstrating the effectiveness of weight-space steering in generalizing to unseen tasks. In conclusion, CreativityNeuro improves divergent thinking and reduces mode collapse without requiring behavioral data, re-training, or gradient-based fine-tuning, providing a straightforward way to enhance LLM performance in creative domains.
\end{abstract}

%% file: sections/03_introduction.tex
\section{Introduction}


Recent advances in large language models (LLMs) have renewed interest in a longstanding question: \emph{how can we understand and enhance creativity in intelligent systems?} \citep{Boden2004TheMechanisms}. While this question has deep roots in cognitive science \citep{a_treatise_on_man, galton_hereditary_genius, Hadamard1954AnField, Guilford1956PsychologicalINTELLECT, Mednick1962TheProcess., Koestler1964TheCreation, Simonton2004CreativityZeitgeist, ArneDietrich2004TheCreativity, Fauconnier2008TheComplexities, Rothenberg2014FlightCreativity}, it is now increasingly studied in the context of large-scale generative models \citep{Maher2010EvaluatingSystems, Varshney2019MathematicalCreativity, Schapiro2025TransformationalTheory}.
Recent work has begun to assess the capacity for LLMs to engage in creative and open-ended tasks \citep{Si2024CanResearchers,Si2025TheIdeas,Sanyal2025Spark:Generation, Bellemare-Pepin2024DivergentLLMs,Wang2025LargeIdeas, Wang2024SciMON:Novelty}, where a recurring issue has surfaced: models tend to consistently generate similar responses to open-ended questions, in what has been termed the \emph{artificial hivemind} effect \citep{hivemind}.

Within the creativity literature, a common distinction is made between \emph{divergent thinking}, the capacity to generate multiple diverse solutions to a problem, and \emph{convergent thinking}, the ability to find a single correct solution that unifies multiple diverse stimuli \citep{Dietrich2019TypesCreativity, Guilford1956PsychologicalINTELLECT}. Studying ways to enhance divergent thinking offers a promising pathway to encourage greater diversity and novelty in model responses, combating the homogenization issues that have emerged thus far. Here, we introduce a weight-space steering method that improves divergent thinking in LLMs. Our method outperforms prior approaches---including decoding, prompting, and activation steering---and generalizes better to unseen tasks, without requiring behavioral data or gradient-based fine-tuning. In detail, our main contributions are as follows:

\begin{enumerate}
    \item In~\Cref{sec:method}, we introduce CreativityNeuro, a data-free method for steering creative behavior. 
    \item In \Cref{sec:dat}, we find that CreativityNeuro significantly improves divergent thinking on the Divergent Association Task (DAT), outperforming baselines such as prompting, activation steering, and decoding baselines.
    \item In~\Cref{sec:aut_tt}, we conduct a large-scale human evaluation on the Alternative Uses Test (AUT) and Task Task (TT) and find that CreativityNeuro improves originality, surprise, and creativity on the AUT and TT, while activation steering transfers poorly to the AUT and TT.
    \item In~\Cref{sec:mode_collapse}, we find that CreativityNeuro reduces mode collapse across all three tasks.
    \item In~\Cref{sec:dt_vs_factual}, we find evidence that divergent thinking and factual reasoning are non-separable in weight space.
\end{enumerate}

%% file: sections/04_background.tex
\section{Related Work}
We start by briefly reviewing related work before introducing our method in~\Cref{sec:method}.

\textbf{Evaluating the creativity of LLMs.} Previous work has evaluated LLMs on divergent creativity assessments--including the DAT \citep{Olson2021NamingCreativity}, AUT \citep{Guilford1956PsychologicalINTELLECT}, the Task Task \citep{chu2024task}--and in various real-world settings like scientific ideation \citep{Si2024CanResearchers, Si2025TheIdeas} and open-ended user queries \citep{hivemind}. On the DAT, LLMs can achieve scores well into the 90th percentile of humans \citep{Bellemare-Pepin2024DivergentLLMs, Wang2025LargeIdeas}, whereas \citet{Stevenson2022PuttingTest} studied GPT-3 on the AUT and concluded that humans  exhibited greater creativity, with model responses showing weaker originality. Lastly, \citet{chu2024task} found that model-generated goals on the Task Task achieved similar creativity ratings as human-generated goals, as assessed by a large panel of human raters.

\textbf{Improving the creativity of LLMs.} Most similar to this work, \citet{Olson2024SteeringCreativity} proposed an activation steering method to amplify the creativity of LLMs, although improvements were only established for a single model, task, and human annotator. Our study is the first to demonstrate a steering method to improve creative behavior and validate its effectiveness in a large-scale human study. Apart from steering, various other approaches to improving LLM creativity include prompting frameworks \citep{dc_framework, Morain2025PromptKnob, Wang2025LargeIdeas}, varying decoding parameters such as temperature \citep{Peeperkorn2024IsModels}, and reinforcement learning (RL) on preference data \citep{rlaif_creativity}. Unlike steering and RL-based approaches that typically require labeled behavioral data, our method operates entirely data-free.

%% file: sections/05_method.tex
\input{tables/cn_algo}
\input{tables/prompt_sets.tex}

\section{Method} \label{sec:method}
Recently, \citet{Christ2025MathPasses} has shown that parameter importance methods can be used to identify and amplify weights involved in mathematical reasoning, improving scores on the MATH benchmark by 4--17\% \citep{Hendrycks2021MeasuringDataset}. Unlike mathematical reasoning, which can be elicited and evaluated on structured benchmarks such as MATH and GSM8K, creativity is a property of \emph{responses}, not of questions. Open-ended prompts, such as those in \citet{hivemind}, admit a wide range of potentially creative completions, but novelty and usefulness are measured on the outputs \citep{Varshney2019MathematicalCreativity, Maher2010EvaluatingSystems, Boden2004TheMechanisms} rather than the inputs themselves. Therefore, our key methodological innovation is a framework for extending MathNeuro to a cognitive domain where the target behavior can be prompted but no structured dataset exists, making CreativityNeuro entirely data-free.

\textbf{Contrastive Prompt Sets} \quad MathNeuro relies on questions drawn from MATH and GSM8K to obtain inputs for parameter importance scoring. Because no analogous dataset exists for creativity, we instead construct \emph{contrastive prompt sets}: short instructions that direct the model toward creative ($\mathcal{P}^\text{cre}$) versus non-creative ($\mathcal{P}^\text{non-cre}$) behavior. We use six such sets spanning a range of styles---\textsc{dat}, \textsc{storytelling}, \textsc{ideation}, \textsc{problem solving}, \textsc{open-ended}, and \textsc{minimal}---where \textsc{minimal} contains only two- to five-word instructions (e.g., \emph{Surprise me} vs \emph{Be precise}). Representative examples are given in~\Cref{tab:prompt_sets}, and all six prompt sets are given in~\Cref{tab:full_prompt_sets}. As a result, CreativityNeuro does not require datasets, behavioral generations, scored responses, or labeled examples, making it entirely data-free, unlike \citet{Christ2025MathPasses}.

\textbf{Parameter Importance Scoring} \quad We use the same  Wanda-style \citep{wanda} parameter importance scoring as \citet{Christ2025MathPasses}, restated here for completeness.
This is done by taking the product of weight magnitude and activation norm, summed across a set of prompts $b$ and their token positions $t$: 
$S_{\ell,ij} = \sum_{b,t} |W_{\ell,ij}| \cdot \|\mathbf{x}^{(b,t)}_{\ell,j}\|_2$, 
where $\mathbf{x}^{(b,t)}_{\ell,j}$ is the $j$-th input activation at layer $\ell$ for token $t$ in prompt $b$. We compute importance scores on creative $\mathcal{P}^\text{cre}$ and non-creative prompts $\mathcal{P}^\text{non-cre}$. Then, we isolate creativity-specific parameters by selecting the top $\rho$ percent of weights ranked by creative importance that do not also appear in the top $\rho$ percent for non-creative prompts. This set difference operation ($C_\ell \setminus N_\ell$) ensures we identify parameters uniquely associated with creative behavior. At inference time, we multiply weights in $C_\ell \setminus N_\ell$ by a scaling factor $(1 + \alpha)$. The hyperparameters $\rho$ and $\alpha$  control the importance threshold and scaling strength, respectively. The full procedure is given in \Cref{alg:creativityneuro}. 

%% file: tables/cn_algo.tex
\begin{algorithm}[t]
\small
\caption{CreativityNeuro}
\label{alg:creativityneuro}
\begin{algorithmic}[1]
\REQUIRE Model weights $\{W_\ell\}_{\ell=1}^L$; creative prompts $\mathcal{P}^{\text{cre}}$; non-creative prompts $\mathcal{P}^{\text{non-cre}}$; importance threshold $\rho$; scaling factor $\alpha$
\ENSURE Modified weights $\{W'_\ell\}_{\ell=1}^L$ with amplified creativity weights
\STATE 
\FOR{each layer $\ell$} 
    \STATE Run forward passes on $\mathcal{P}^{\text{cre}}, \mathcal{P}^{\text{on-cre}}$ to obtain importance scores $S^{\text{cre}}_{\ell,ij}, S^{\text{non-cre}}_{\ell,ij}$ via:
    \begin{align*} \label{eq:cn}
        S_{\ell,ij}(\mathcal{P}) = \sum_{b=1}^{|\mathcal{P}|} \sum_{t=1}^{T_b} |W_{\ell,ij}| \cdot \|\mathbf{x}^{(b,t)}_{\ell,j}\|_2 \quad\quad\quad\quad\quad\quad \textbf{Step 1: Compute weight importance scores} 
    \end{align*}
    for prompt $b$ and token position $t$ within prompt $b$  
\ENDFOR
\STATE
\FOR{each layer $\ell$} 
    \STATE $C_\ell \gets$ top-$\rho$ weights ranked by $S^{\text{cre}}_{\ell,ij}$  \hfill  \textbf{Step 2: Extract creative-specific subspaces}
    \STATE $N_\ell \gets$ top-$\rho$ weights ranked by $S^{\text{non-cre}}_{\ell,ij}$
    \STATE $M^{\text{cre-spec}}_{\ell,ij} \gets \mathbb{I}[(i,j) \in C_\ell \setminus N_\ell]$
\ENDFOR
\STATE 
\FOR{each layer $\ell$} 
    \STATE $W'_\ell \gets W_\ell \odot (1 + \alpha \cdot M^{\text{cre-spec}}_\ell)$ \hfill \textbf{Step 3: Creative parameter scaling}
\ENDFOR
\STATE \textbf{return} $\{W'_\ell\}_{\ell=1}^L$
\end{algorithmic}
\end{algorithm}

%% file: tables/prompt_sets.tex
\begin{table*}[tb]
\centering
\small
\caption{\textbf{Examples of contrastive prompt sets.} Each set contains creative ($\mathcal{P}^\text{cre}$) and non-creative ($\mathcal{P}^\text{non-cre}$) prompts used for parameter importance scoring in \Cref{alg:creativityneuro}.}
\label{tab:prompt_sets}
\begin{tabular}{p{0.08\textwidth} p{0.42\textwidth} p{0.42\textwidth}}
\toprule
\textbf{Set} & \textbf{Creative} ($\mathcal{P}^\text{cre}$) & \textbf{Non-Creative} ($\mathcal{P}^\text{non-cre}$) \\
\midrule
\textsc{story}
& \scriptsize Write the first line of a story that makes the reader question reality.  
& \scriptsize Create a typical story beginning that establishes setting and character clearly. \\
\midrule
\textsc{problem}
& \scriptsize What would an alien civilization's approach to this problem look like?
& \scriptsize What best practices should be followed when solving this? \\
\midrule
\textsc{minimal}
& \scriptsize Surprise me.
& \scriptsize Be precise. \\
\midrule
\end{tabular}
\end{table*}

%% file: sections/06_dat.tex
\section{Experiments on the Divergent Association Task} \label{sec:dat}

\begin{figure}[t]
    \centering
    \includegraphics[width=\textwidth]{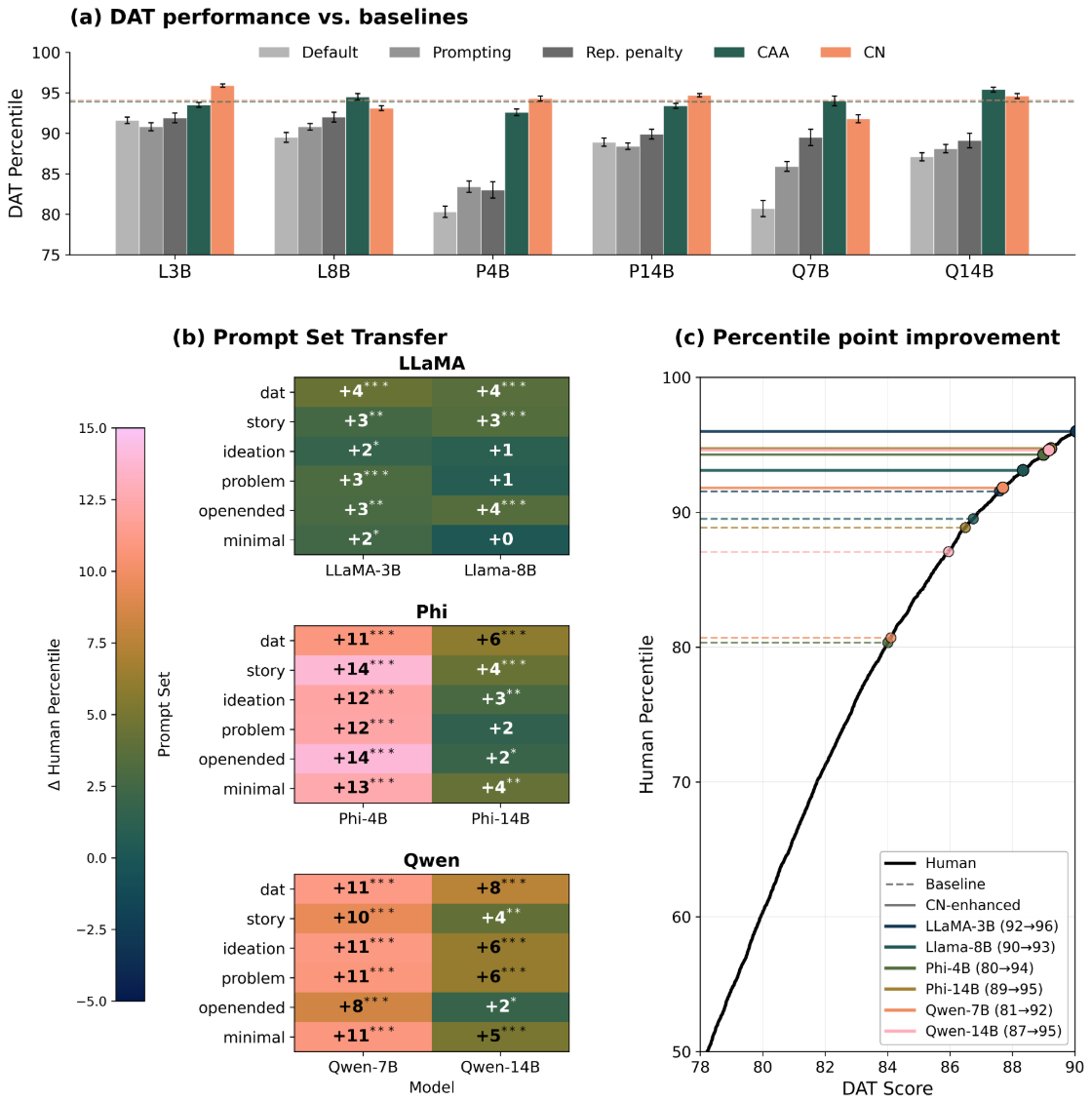}
    \caption{\textbf{CreativityNeuro (CN) improves divergent thinking across models and prompt sets.} Given a human reference distribution \citep{Wang2025LargeIdeas} ($N = 9{,}297$, $\mu = 78.26$, $\sigma = 6.73$), we report:
    (a)~DAT human percentile ($\pm$SEM) averaged across $T \in \{0.9, 1.0, 1.2\}$ for CN, CAA, and the strongest sampling-based baselines; dashed lines show cross-model means for CN and CAA. (b)~Heatmap showing human percentile improvement ($\Delta$\%ile) for CreativityNeuro models across prompt sets, with statistical significance ($p < 0.05$) at each of the temperatures tested (0.9, 1.0, 1.2) denoted by an asterisk. (c)~CDF showing DAT scores for CreativityNeuro models on the best performing prompt set.}
    \label{fig:cn_enhancement}
    \vspace{-10pt}
\end{figure}

We test instruct-tuned models across three open-weight model families (Phi, Llama, Qwen), totaling six models at 3B, 4B, 7B, 8B, and 14B sizes: LLaMA (3.2-3B-Instruct, 3.1-8B-Instruct) \citep{grattafiori2024llama3herdmodels}, Qwen-2.5 (7B-Instruct, 14B-Instruct) \citep{qwen_technical_report}, and Phi (3.5-mini-Instruct (4B), 3-medium-4k-Instruct (14B)) from Microsoft \citep{phi_technical_report}. For each model we use the best $(\rho, \alpha, \text{prompt set})$ configuration with $\geq 120$ valid CN samples at all three temperatures $T \in \{0.9, 1.0, 1.2\}$. Full hyperparameter sweep details are in \Cref{sec:hyperparam_sweeps}.

\textbf{Task} \quad The DAT asks participants to generate $N$ words that are as semantically distant from each other as possible \citep{Olson2021NamingCreativity}. Given a set of $N$ words $W := \{w_1, w_2, \dots, w_N\}$ with corresponding GloVe embeddings $V := \{\mathbf{v}_1, \mathbf{v}_2, \dots, \mathbf{v}_N\} \subseteq \mathbb{R}^{300}$, the DAT score is the average pairwise semantic distance among all distinct pairs of those $N$ words:
\begin{equation}
    \textrm{DAT}(W) := \frac{100}{N(N-1)} \sum_{i \neq j}^N (1 - \cos(\mathbf{v}_i, \mathbf{v}_j))
\end{equation}
Following \citet{Olson2021NamingCreativity}, we use the 840B-token GloVe embeddings \citep{Pennington2014GloVe} as our semantic space. A participant is asked to name $N=10$ words, and the first $7$ valid words are kept \citep{Olson2021NamingCreativity}. Full prompts are given in~\Cref{sec:eval_prompts}.

\textbf{Baselines} \quad \label{par:dat_baselines} Existing studies have found that temperature-scaling and prompting can influence DAT scores \citep{Wang2025LargeIdeas, Bellemare-Pepin2024DivergentLLMs}. To ensure the CN intervention leads to a meaningful improvement over such techniques, we compare against a broad set of baselines, including prompting; varying decoding parameters such as top-p nucleus sampling, top-k sampling, temperature, and repetition penalty; as well as activation steering via contrastive activation addition (CAA; \citet{Panickssery2024SteeringVectors}), which injects a steering vector $\mathbf{v}_\ell = \bar{\mathbf{h}}_\ell^{+} - \bar{\mathbf{h}}_\ell^{-}$ into the residual stream during decoding. For activation steering, contrast pairs are obtained from top- vs.\ bottom-quartile DAT responses by score, creating a ``divergent thinking'' direction in the residual stream.\footnote{We also tested a prompt-only CAA variant using forward-pass activations from the same creative vs.\ non-creative prompt sets as CN, but it underperformed the behavioral-data variant on all models and is omitted for clarity.} Full decoding strategy and activation steering hyperparameter settings are given in \Cref{sec:baseline_sweep_settings}. All settings are evaluated across all six models at $T \in \{0.9, 1.0, 1.2\}$ until $N{=}120$ valid DAT responses are obtained.

\subsection{Results}

\textbf{CreativityNeuro achieves robust performance improvements across prompt sets} \quad
CN improves DAT performance across all six models and prompt sets (\Cref{fig:cn_enhancement}), outperforming all sampling-based baselines. Panel~(b) of \Cref{fig:cn_enhancement} shows $\Delta$\,Percentile for the best $(\rho, \alpha)$ per model--prompt combination: while the \textsc{dat} prompt set produces the most consistent gains, non-\textsc{dat} prompt sets also yield statistically significant improvements, suggesting CN is able to identify weights controlling divergent thinking behavior, rather than localizing DAT-specific task knowledge.

\textbf{CreativityNeuro outperforms activation steering without needing behavioral data} \quad
CN (94.1 avg) slightly outperforms activation steering (93.9 avg) on DAT percentile (\Cref{fig:cn_enhancement}a); however, activation steering requires scored DAT responses to construct its steering vector, while CN uses only creative and non-creative prompts with no generation or scoring. Prompt-only activation steering (omitted from the figure; see footnote in~\Cref{par:dat_baselines}) averaged only 87.8, comparable to prompting (87.9), suggesting that the behavioral data is essential for activation steering to be competitive, whereas CN achieves stronger performance from prompts alone.

\begin{takeaway}
CreativityNeuro improves DAT scores across all six models and prompt sets, outperforming prompting, sampling-parameter, and activation steering baselines.
\end{takeaway}

%% file: sections/07_aut_tt.tex
\section{Experiments on the Alternative Uses Test and Task Task} \label{sec:aut_tt}
Here, we evaluate CreativityNeuro on more complex divergent thinking tasks than the DAT. The Alternative Uses Test (AUT), a standard instrument in the psychometrics literature, asks participants to generate creative uses for a common object \citep{Guilford1956PsychologicalINTELLECT}. We administer the AUT using a standard set of objects in the creativity literature: \emph{brick}, \emph{paperclip}, and \emph{fork}. Following \citet{Stevenson2022PuttingTest}, uses are scored on \emph{originality}, \emph{surprise}, and \emph{utility}. We also evaluate CreativityNeuro on the Task Task (TT), which assesses the ability to generate novel challenges or goals that themselves expect novel solutions \citep{chu2024task}. Participants design creative game show challenges that would be fun to attempt, entertaining to watch, and difficult enough to be interesting. Following \citet{chu2024task}, we evaluate responses on creativity and originality.\footnote{\citet{chu2024task} studies additional dimensions, such as difficulty, how fun the task is to do, and how fun it is to watch, but here we restrict focus to creativity and originality, as these are most relevant to the goal of evaluating divergent thinking.} Full prompts are in~\Cref{sec:eval_prompts}.

\begin{figure*}[t]
    \centering
    \includegraphics[width=0.9\textwidth]{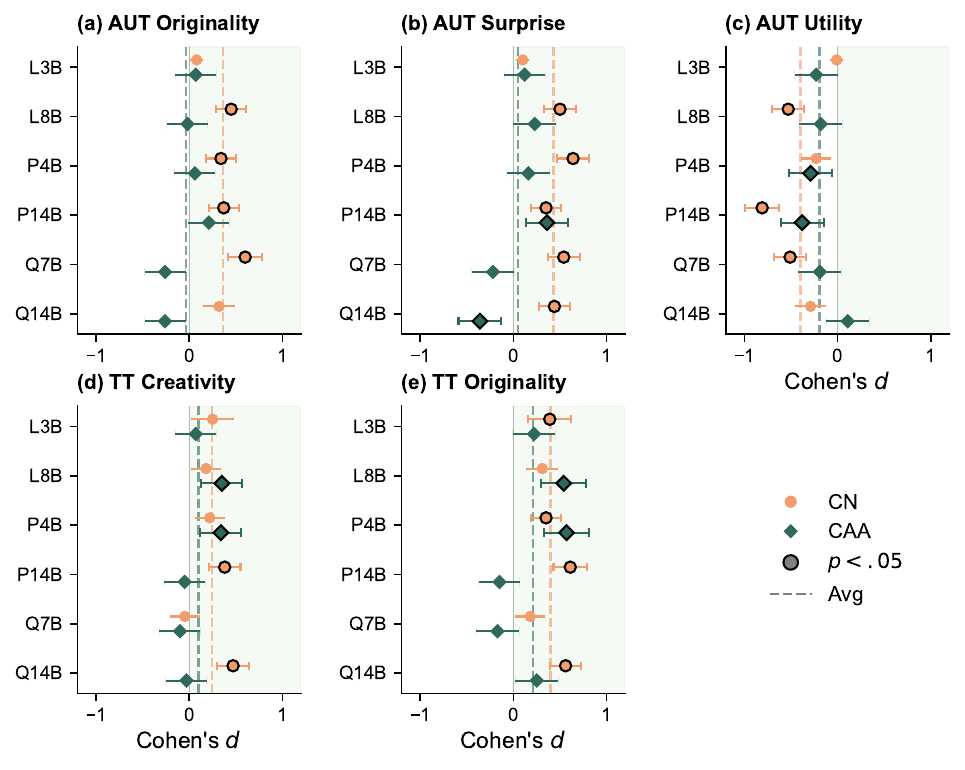}
    \caption{\textbf{Cohen's $d$ ($\pm$SE) from intra-participant $z$-scored human ratings on the AUT and TT.} \textbf{(a--c)}~AUT Originality, Surprise, Utility. \textbf{(d--e)}~TT Creativity, Originality. Black outlines indicate $p < .05$, and green shading marks the positive-effect region.}
    \label{fig:results}
\end{figure*}

\textbf{Models and Inference} \quad For each of the six models in~\Cref{sec:dat}, we select the CreativityNeuro configuration ($\rho$, $\alpha$, prompt set) that produced the largest statistically significant DAT improvement. Then, for each task, we sample 40 stimuli (20~baseline, 20~creative) at temperature $T=1.0$, $\text{top-}k = 0$, and $\text{top-}p = 1.0$. We compare CreativityNeuro against activation steering, the strongest non-CN baseline from~\Cref{sec:dat}. 

\textbf{Human Experiment Design} \quad We evaluate AUT and TT stimuli via human ratings on Prolific. Human studies use a between-subjects design, where every participant rates 10~stimuli (5~baseline, 5~creative) in randomized order with condition labels hidden. We ensure balanced allocation via automated participant-to-slot assignment so that each stimulus receives exactly 10~independent ratings. Ratings on the AUT and TT are given on continuous 0--100 sliders, and to control for individual differences in scale usage, we compute intra-participant $z$-scores: for each participant $i$ and dimension $d$, $z_{i,d,s} = (r_{i,d,s} - \bar{r}_{i,d})\,/\,\sigma_{i,d}$, where $\bar{r}_{i,d}$ and $\sigma_{i,d}$ are computed across all stimuli that participant rated on that dimension. We recruit 30 participants per (model, task, method) triple, reaching N=720 total human reviewers. Effect sizes and significance ($t$-tests\footnote{We confirm responses follow a normal distribution before applying the $t$-tests.}) are computed on $z$-scored ratings for baseline vs.\ creative, where each participant is treated as an independent sample.

\input{figures/cn_stimuli_examples}

\subsection{Results}

\textbf{CreativityNeuro improves originality, surprise, and creativity} \quad We report results in~\Cref{fig:results}. On the AUT, CreativityNeuro achieves uniformly positive originality effects across all six models (avg.\ $d = +.36$), with four reaching significance. The effects are even stronger for surprise (avg.\ $d = +.43$), with five models reaching significance. On the Task Task, CreativityNeuro obtains strong originality gains (avg.\ $d = +.40$), with the strongest effects in Phi-14B ($d = +.61$) and Qwen-14B ($d = +.56$), and moderate improvements in overall creativity (avg.\ $d = +.24$). Although CreativityNeuro degrades AUT utility, this is a predictable consequence of the novelty--utility tradeoff already present in baseline responses. See Appendix~\Cref{sec:novelty_utility} for detailed analysis of this tradeoff.

\textbf{CreativityNeuro generalizes better to the AUT and TT than activation steering} \quad Additionally, while activation steering (93.9 avg) performs comparably to CreativityNeuro (94.1 avg) on the DAT (\Cref{fig:cn_enhancement}a), activation steering fails to generalize to the AUT and Task Task. These results are consistent with recent work that has found weight-space steering generalizes further out-of-distribution than activation steering on sycophancy and value alignment tasks \citep{fierro2025steering}. Moreover, activation steering effectiveness has been shown to vary significantly by behavior type, with more complex behaviors like embodying persona archetypes and public figures proving more difficult to steer \citep{limits_of_as}. Techniques such as context-dependent \citep{context_dependent_steering, conditional_as} and learned activation steering \citep{learned_activation_steering} have been proposed to remediate such issues, and may be explored in future work.

\begin{takeaway}
CreativityNeuro generalizes to open-ended creative tasks judged by human raters, improving measures of originality and surprise, while activation steering does not exhibit reliable transfer.
\end{takeaway}

%% file: figures/cn_stimuli_examples.tex
\begin{figure*}[t]
\centering
\small
\setlength{\fboxsep}{6pt}
\setlength{\fboxrule}{0.6pt}
\definecolor{batlowcool}{HTML}{1C5A61}
\definecolor{batlowwarm}{HTML}{E8985C}

(a) \textbf{Alternative Uses Test} --- ``List alternative uses for a \underline{\phantom{object}}.''
\vspace{4pt}

\begin{minipage}[t]{0.48\textwidth}
\colorbox{batlowcool!10}{\begin{minipage}{\dimexpr\linewidth-2\fboxsep}
\textbf{\color{batlowcool!80!black}CN-enhanced (Llama-8B), $\bar{z}=0.67$}\\[2pt]
\textbf{Book:} A book can be transformed into a giant marimba by hollowing out its pages, gluing wood strips to the hollowed-out sections, and tuning metal or glass material to create a makeshift percussion instrument.\\[1pt]
{\tiny\color{gray} originality\,=\,1.12 \quad surprise\,=\,1.32 \quad utility\,=\,$-$0.42}
\end{minipage}}
\vspace{4pt}

\end{minipage}%
\hfill
\begin{minipage}[t]{0.48\textwidth}

\colorbox{gray!8}{\begin{minipage}{\dimexpr\linewidth-2\fboxsep}
\textbf{\color{gray!60!black}Baseline (Phi-4B), $\bar{z}=0.36$}\\[2pt]
\textbf{Book:} A book can transform into a unique, eco-friendly planter, nurturing a tiny ecosystem as it blossoms with flowers.\\[1pt]
{\tiny\color{gray} originality\,=\,0.58 \quad surprise\,=\,0.70 \quad utility\,=\,$-$0.20}
\end{minipage}}
\end{minipage}

\vspace{8pt}
(b) \textbf{Task Task} --- ``Design a creative game show challenge.''
\vspace{4pt}

\begin{minipage}[t]{0.48\textwidth}
\colorbox{batlowwarm!10}{\begin{minipage}{\dimexpr\linewidth-2\fboxsep}
\textbf{\color{batlowwarm!100!black}CN-enhanced (Qwen-14B), $\bar{z}=1.07$}\\[2pt]
Your objective is to construct a functional periscope allowing clear line-of-sight around an obstacle, mimicking an espionage thrill; you must see a small target widget placed behind a bookcase [\ldots] You may utilize a stack of encyclopedias, a roll of wrapping paper, a broomstick, two mirrors, safety goggles, and a length of garden hose... \\[1pt]
{\tiny\color{gray} creativity\,=\,1.08 \quad originality\,=\,1.06}
\end{minipage}}
\end{minipage}%
\hfill
\begin{minipage}[t]{0.48\textwidth}
\colorbox{gray!8}{\begin{minipage}{\dimexpr\linewidth-2\fboxsep}
\textbf{\color{gray!60!black}Baseline (Qwen-14B), $\bar{z}=0.44$}\\[2pt]
Your goal is to create a makeshift catapult using only a skateboard, a bicycle tire inner tube, a cooking pot lid, and six feet of string. Using this device, you must launch a raw egg into an oversized wading pool filled with colorful foam noodles placed fifteen feet away. Each successful catapult that lands the egg in the pool scores five points; however, the egg must remain intact upon impact to earn the full five points.\\[1pt]
{\tiny\color{gray} creativity\,=\,0.50 \quad originality\,=\,0.37}
\end{minipage}}
\end{minipage}

\caption{\textbf{Top-rated CreativityNeuro vs.\ baseline generations from the same model.} Intra-participant $z$-scores averaged across raters ($N{=}30$ per cell). (a) CreativityNeuro responses on the AUT tend to score higher on originality and surprise. (b) CreativityNeuro challenges on the Task Task.}
\label{fig:cn_stimuli_examples}
\end{figure*}

%% file: sections/08_mode_collapse.tex
\begin{figure*}[t]
    \centering
    \includegraphics[width=\textwidth]{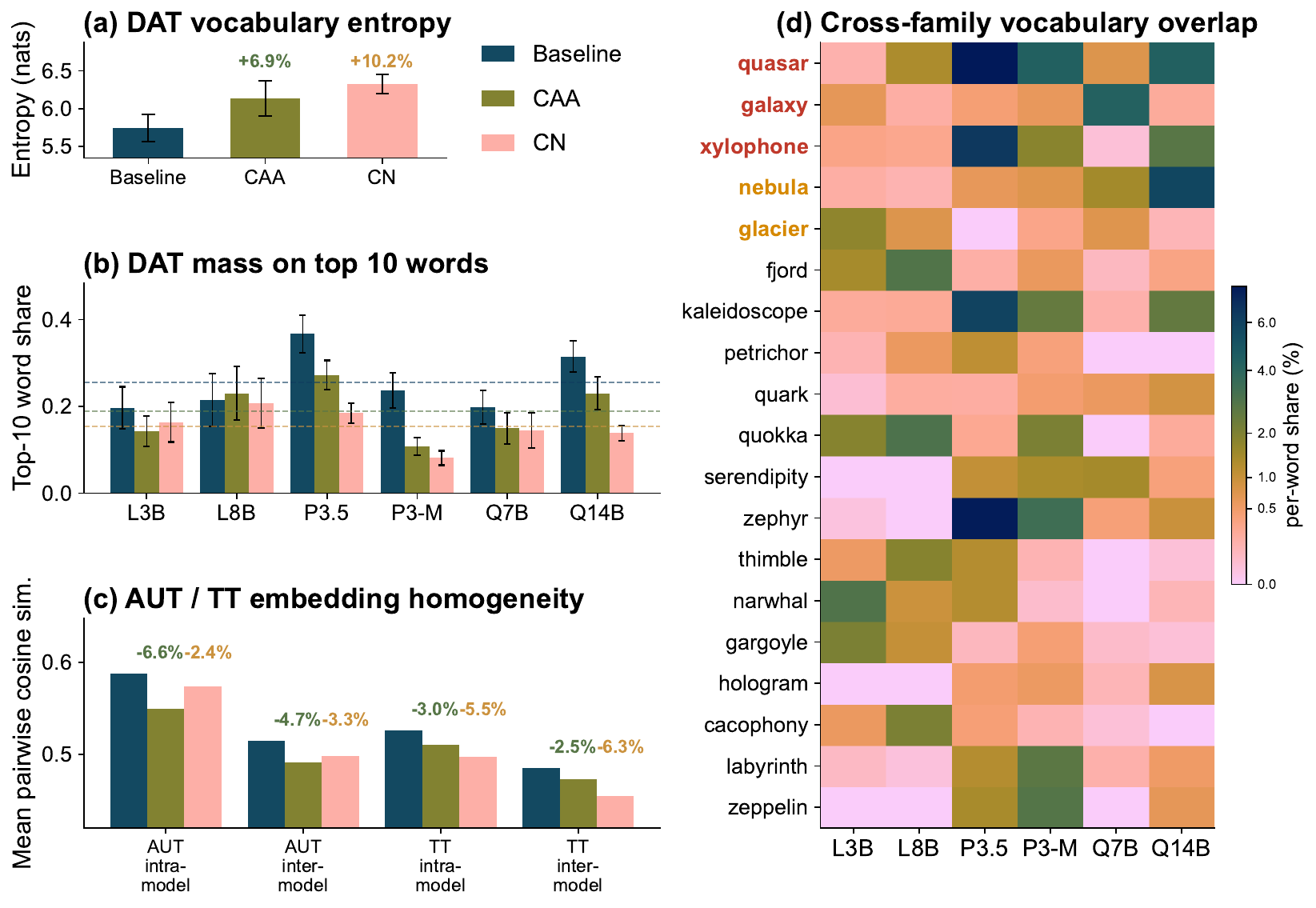}
    \caption{\textbf{Measures of mode collapse across tasks.} Baseline / CAA / CN shown left-to-right (teal / green / orange). \textbf{(a)} DAT vocabulary entropy. \textbf{(b)} DAT top 10 word share. \textbf{(c)} AUT and TT embedding homogeneity. \textbf{(d)} Cross-family vocabulary overlap.}
    \label{fig:mode_collapse}
\end{figure*}

\section{Evaluating Mode Collapse Across Tasks} \label{sec:mode_collapse}
Instruction-tuned LLMs are known to suffer from mode collapse--the tendency to concentrate open-ended outputs on a narrow set of semantic clusters \citep{hivemind, springer_mode_collapse}. In this section, we test whether CreativityNeuro reduces word-level mode collapse on the DAT and response-level collapse on the AUT and TT.

\textbf{Metrics} \quad On the DAT, we pool outputs across temperatures $T \in \{0.9, 1.0, 1.2\}$ and compute each model's vocabulary entropy $H$ and probability mass assigned to each~(model, condition $\in \{$ baseline, CAA, CN $\}$)'s own top 10 most-frequent words. On the AUT and TT, where responses span multiple sentences, we adopt the intra-model repetition (mean pairwise cosine similarity within a model's responses) and inter-model homogeneity (mean pairwise cosine similarity between responses from different models on the same query) metrics from \citet{hivemind}, also using \texttt{openai/text-embedding-3-large} for response embedding.

\subsection{Results}

\textbf{Baseline models suffer significantly from mode collapse on the DAT.} \quad Baseline models concentrate $25.5\%$ of generated tokens on average on each model's top-10 most frequent words (\Cref{fig:mode_collapse}b, averaged across $T \in \{0.9, 1.0, 1.2\}$). Furthermore, $19$ words appear in the top 30 vocabulary of at least three of the six tested models, as shown in \Cref{fig:mode_collapse}d. Three of these words---\textcolor[HTML]{c0392b}{\emph{galaxy, quasar, xylophone}}---appear in the baseline top 30 of \emph{all three model families} (LLaMA, Phi, Qwen) simultaneously, and two more---\textcolor[HTML]{d68900}{\emph{glacier, nebula}}---appear in the baseline top-30 of $\geq 4$ of the six models.

\textbf{CreativityNeuro reduces word-level mode collapse on the DAT.} \quad As shown in \Cref{fig:mode_collapse}b, CreativityNeuro reduces the top 10 share by $10.2$ pp on average and increases vocabulary entropy by $0.59$ nats ($+10\%$, \Cref{fig:mode_collapse}a). Activation steering (CAA) produces a smaller but qualitatively similar effect: top-10 share drops by $6.6$ pp ($0.255 \to 0.189$), and vocabulary entropy increases by $0.40$ nats ($+7\%$).

\textbf{CreativityNeuro and activation steering both reduce mode collapse on the AUT and TT} \quad In~\Cref{fig:mode_collapse}c, we report relative reductions vs.\ baseline annotated above each CAA bar (green) and CN bar (orange). Both CN and CAA reduce homogeneity on each task. On the AUT (sentence-length responses), CAA reduces intra-model repetition by $6.6\%$ and inter-model homogeneity by $4.7\%$, larger than CN's $2.4\%$ and $3.3\%$ respectively. On the TT (paragraph-length responses), CN reduces intra-model repetition by $5.5\%$ and inter-model homogeneity by $6.3\%$, larger than CAA's $3.0\%$ and $2.5\%$.

\begin{takeaway}
CreativityNeuro reduces semantic mode collapse on both word-level (DAT) and embedding-level (AUT, TT) measures. Activation steering produces a similar but smaller effect on the DAT and TT, while matching CreativityNeuro on the AUT.
\end{takeaway}

%% file: sections/09_dt_vs_factual.tex
\section{Are Divergent Thinking and Factual Reasoning Separable in Weights?} \label{sec:dt_vs_factual}
Previously, \citet{Christ2025MathPasses} demonstrated that MathNeuro could improve MATH benchmark scores without degrading factual reasoning on the Massive Multi-task Language Understanding (MMLU) benchmark \citep{Hendrycks2021MeasuringDataset}. Here, we study whether improved divergent thinking interferes with factual reasoning on MMLU.

\textbf{Setup} \quad For each of the six models, we take the top performing $(\rho, \alpha$, prompt) configuration from \Cref{sec:dat} and measure 5-shot accuracy on 500 MMLU questions, across 5 seeds per model, under two mask construction techniques:
\begin{enumerate}
    \item \textbf{Default:} $P^\text{cre} \setminus P^\text{non-cre}$ uses the default creative and non-creative prompts from \Cref{sec:dat}.
    \item \textbf{MMLU-protected:} $P^\text{cre} \setminus (P^\text{non-cre} \cup P^\text{MMLU})$ adds 20 randomly sampled MMLU prompts $\mathcal{P}^\text{MMLU}$ to the negative contrast set in \Cref{alg:creativityneuro}, further removing any weight whose importance is above the $1-\rho$ percentile for MMLU.
\end{enumerate}
In summary, the MMLU-protected masks attempts to explicitly separate creative weights from MMLU weights. If divergent thinking and factual reasoning are fully separable, the MMLU-protected mask should preserve MMLU scores without degrading DAT scores.

\subsection{Results}

\begin{wrapfigure}[22]{r}{0.45\textwidth}
    \vspace{-12pt}
    \centering
    \includegraphics[width=0.45\textwidth]{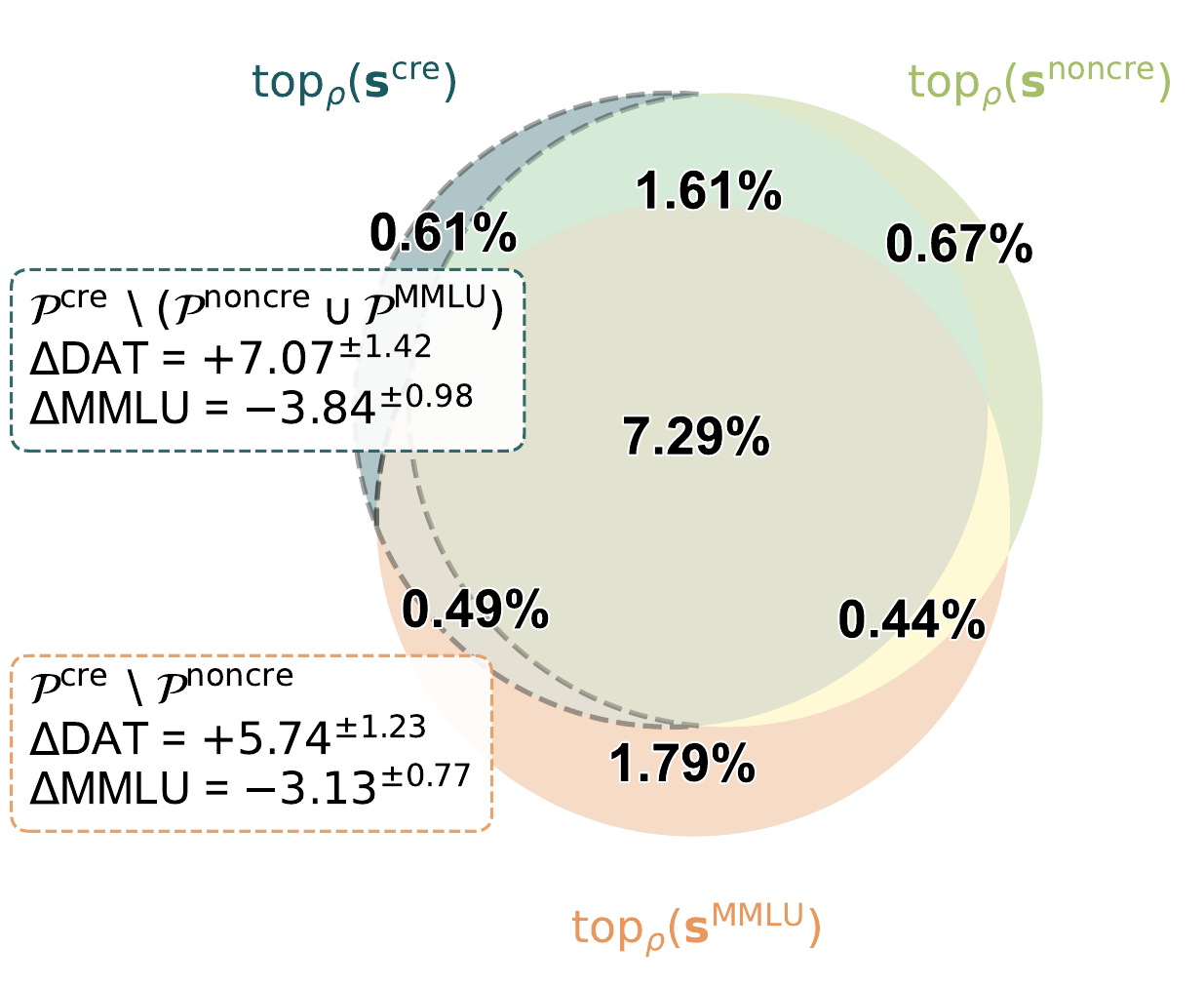}
    \caption{\textbf{Parameter importance across masks} at $\rho{=}0.1$. The \textcolor[HTML]{E8985C}{default mask $\mathcal{P}^{\mathrm{cre}} \setminus \mathcal{P}^{\mathrm{noncre}}$} and the \textcolor[HTML]{1C5A61}{MMLU-protected mask $\mathcal{P}^{\mathrm{cre}} \setminus (\mathcal{P}^{\mathrm{noncre}} \cup \mathcal{P}^{\mathrm{MMLU}})$} are the dotted regions on the left. Annotated $\Delta$DAT (percentile) and $\Delta$MMLU (pp) are cross-model means $\pm$ SEM.}
    \label{fig:mmlu_augmentation}
    \vspace{-5pt}
\end{wrapfigure}

\textbf{Divergent thinking and factual reasoning are non-separable in weight space} \quad Default masks reduce MMLU accuracy by $-3.13$ pp on average (\Cref{fig:mmlu_augmentation}). Meanwhile, MMLU-protected masks gain an additional $+1.33$ percentile $\Delta$DAT on top of default masks, but further reduce MMLU scores by $-0.71$ pp (\Cref{fig:mmlu_augmentation}). Surprisingly, adding MMLU prompts to the negative contrast set \emph{further degrades} MMLU accuracy, despite reducing the size of the masks by $\sim 2 \times$, providing evidence that divergent thinking and factual reasoning are functionally entangled and non-separable in weight space.

Our results are consistent with broader findings in the mechanistic interpretability literature. Namely, it has been shown that individual weights can be entangled in multiple distinct functions (a phenomenon known as \emph{polysemanticity}), which supports the ability of neural models to represent more features than they have neurons (\emph{superposition}) \citep{Elhage2022ToyModels, open_problems_in_mech_interp}. Creativity research more broadly suggests a need for separation between generation and selection \citep{Varshney2019MathematicalCreativity}.
The success of multi-agent creative systems \citep{lin2025creativity} and multi-stage prompting techniques that decouple creative exploration from constraint satisfaction \citep{dc_framework} can be interpreted within the context of these results: if weights controlling divergent and convergent thinking are entangled, simultaneously eliciting strong divergent \emph{and} convergent abilities may be challenging or even impossible. Therefore, separating these steps across agents or prompts can provide stronger overall performance.

\begin{takeaway}
We find evidence, consistent with the broader mechanistic interpretability literature, that divergent thinking and factual reasoning are functionally entangled in model weights.
\end{takeaway}

%% file: sections/10_limitations.tex
\section{Limitations and Future Work} \label{sec:limitations}

Our evaluation is restricted to a finite set of divergent thinking benchmarks and metrics, which capture only certain aspects of creativity. As \citet{Runco2008Commentary:Creativity} notes, divergent thinking is not synonymous with creativity and should best be thought of as a measure of creative potential. Our comparison to activation steering focuses on a standard CAA-style method and does not exhaust the space of possible activation-space interventions, such as context-dependent or learned approaches. Lastly, we found evidence suggesting divergent thinking and factual reasoning are non-separable in model weights---understanding whether this entanglement reflects a fundamental architectural constraint or arises as an artifact of the Wanda-style importance technique remains an important open question. Designing architectures that learn unified factored representations \citep{fer_paper} and disentangle divergent thinking from factual reasoning would be a valuable and timely direction for future work.

%% file: sections/11_appendix.tex
\newpage
\appendix


\section{Full Sets of Contrastive Prompts} \label{sec:full_prompts}

Each of the six prompt sets contains 10 creative and 10 non-creative exemplars. Below we show 3 representative examples from each set.

\begin{table*}[h]
\centering
\small
\caption{\textbf{Full contrastive prompt sets (3 examples each).} Each prompt set contains 10 creative ($\mathcal{P}^\text{cre}$) and 10 non-creative ($\mathcal{P}^\text{non-cre}$) exemplars used for parameter importance scoring in \Cref{alg:creativityneuro}.}
\label{tab:full_prompt_sets}
\begin{tabular}{p{0.08\textwidth} p{0.42\textwidth} p{0.42\textwidth}}
\toprule
\textbf{Set} & \textbf{Creative} ($\mathcal{P}^\text{cre}$) & \textbf{Non-Creative} ($\mathcal{P}^\text{non-cre}$) \\
\midrule
\textsc{dat}
& \scriptsize List 10 common English nouns that are as unrelated in meaning as possible. Avoid any shared topic or category. Output only the nouns, separated by commas.
& \scriptsize List 10 common English nouns that are as closely related in meaning as possible and clearly fit into a single narrow topic. Output only the nouns, separated by commas. \\
& \scriptsize Give me 10 one-word English nouns that are extremely far apart in semantic meaning. They should not fit into a single theme.
& \scriptsize Give me 10 one-word English nouns that are very strongly associated with each other and belong to the same specific domain. \\
& \scriptsize Produce 10 everyday English nouns that share no obvious connection with one another. Each noun should come from a very different domain.
& \scriptsize Produce 10 English nouns that are tightly connected around a single clear theme (for example, all parts of a computer or all items in a kitchen). \\
\midrule
\textsc{story}
& \scriptsize Write an unusual opening sentence for a short story that subverts reader expectations.
& \scriptsize Write a standard opening sentence for a fairy tale. \\
& \scriptsize Create a story beginning that combines two unrelated concepts in a surprising way.
& \scriptsize Create a typical story beginning that establishes setting and character clearly. \\
& \scriptsize Write the first line of a story that makes the reader question reality.
& \scriptsize Write the first line of a conventional mystery novel. \\
\midrule
\textsc{ideation}
& \scriptsize List 5 unusual uses for a brick that nobody has thought of before.
& \scriptsize List 5 common uses for a brick in construction. \\
& \scriptsize Generate unconventional solutions to reduce traffic in cities.
& \scriptsize Generate standard solutions to reduce traffic in cities. \\
& \scriptsize What are some surprising ways a library could be repurposed?
& \scriptsize What are the traditional functions of a library? \\
\midrule
\textsc{problem}
& \scriptsize How might you solve this problem in a way that seems counterintuitive at first?
& \scriptsize What is the most efficient way to solve this problem? \\
& \scriptsize What would an alien civilization's approach to this problem look like?
& \scriptsize List the standard steps for addressing this type of issue. \\
& \scriptsize If you had to solve this with resources from a different era, what would you do?
& \scriptsize What best practices should be followed when solving this? \\
\midrule
\textsc{open}
& \scriptsize Generate a joke about electric vehicles.
& \scriptsize Explain nuclear fission like I am five years old. \\
& \scriptsize Create the first verse of a wedding vow.
& \scriptsize What is Bukhara? Provide a paragraph-long explanation in layman's language. \\
& \scriptsize Write a song about a guy named Jacob working at a call center making jokes.
& \scriptsize In a few sentences explain what threats do scams pose to individuals? \\
\midrule
\textsc{minimal}
& \scriptsize Invent something.
& \scriptsize State a fact. \\
& \scriptsize Surprise me.
& \scriptsize Be accurate. \\
& \scriptsize Be weird.
& \scriptsize Be precise. \\
\bottomrule
\end{tabular}
\end{table*}

\section{Baseline Sweep Settings} \label{sec:baseline_sweep_settings}
In~\Cref{sec:dat}, we compare CreativityNeuro against a set of baseline techniques. 
\begin{enumerate}
    \item \textbf{Prompting:} We present creative exemplars from each of six prompt sets (\Cref{tab:full_prompt_sets}) as in-context guides
    \item \textbf{Decoding Parameter Sweeps}
    \begin{enumerate}
        \item \textbf{top-$p$} nucleus sampling, $p \in \{0.8, 0.85, 0.9, 0.95, 1.0\}$
        \item \textbf{top-$k$} sampling, $k \in \{10, 25, 50, 100, \text{disabled}\}$
        \item \textbf{repetition penalty}, $\theta \in \{1.0, 1.1, 1.2, 1.5, 2.0, 3.0\}$
    \end{enumerate}
    \item \textbf{Activation Steering}: We sweep the injection layer suffix from single-layer to the final 50\% of layers and find that injecting into the final 30\% works best. We sweep $\alpha \in \{0.1, 0.2, 0.3, 0.4, 0.5, 1.0, 2.0, 4.0\}$, selecting the best $\alpha$ where all three temperatures yield $\geq 120$ valid samples.
\end{enumerate}

\section{Layerwise Ablation Studies}

\subsection{Suffix vs.\ Prefix} \label{subsec:layerwise_ablation}

\begin{figure}[t]
    \centering
    \includegraphics[width=\textwidth]{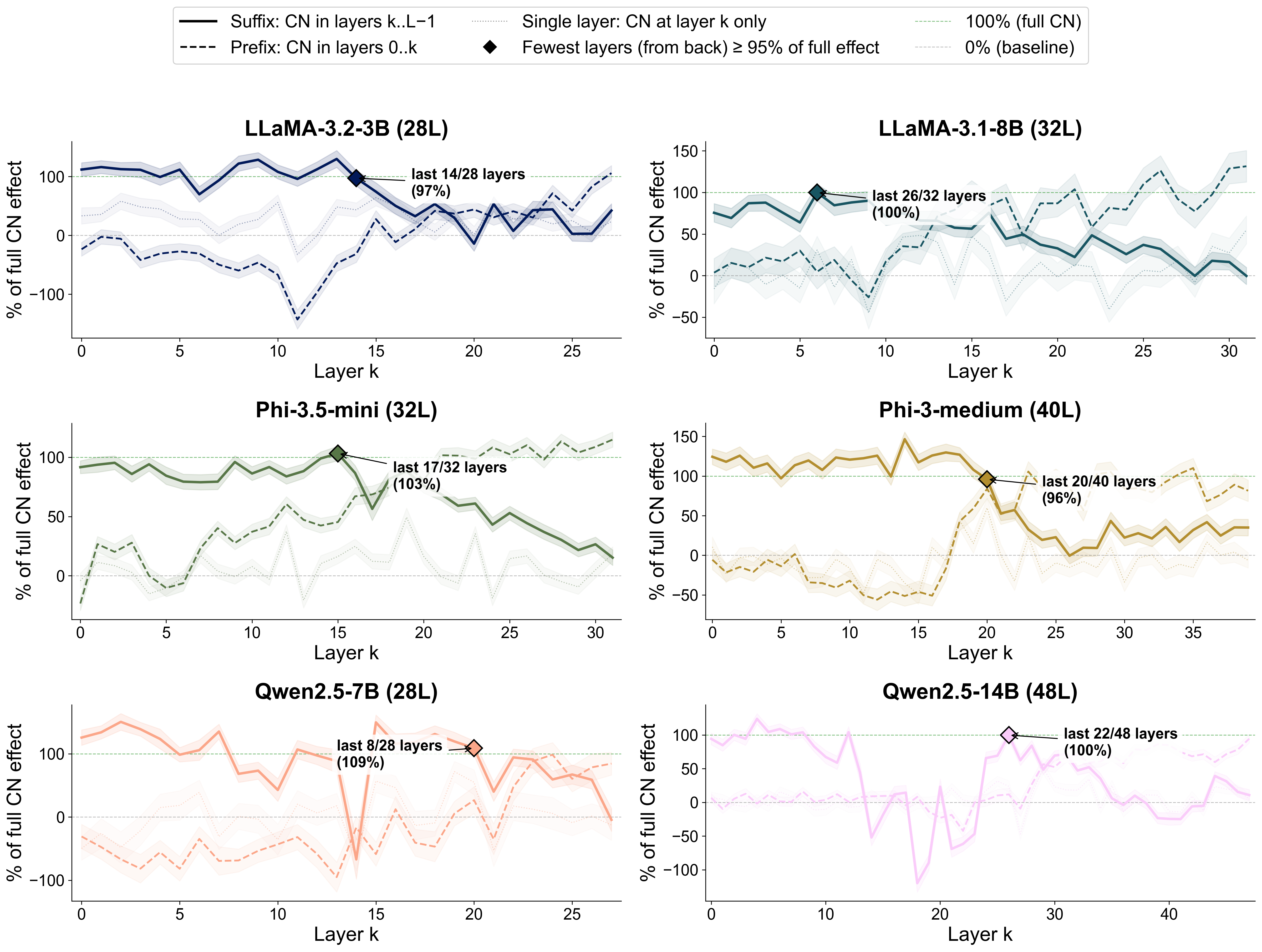}
    \caption{\textbf{Layerwise ablation: suffix vs.\ prefix vs.\ single-layer.} Each panel shows the \% of full CN DAT effect recovered as a function of the number of layers $k$ with CN weights applied. Solid lines: suffix (last $k$ layers). Dashed lines: prefix (first $k$ layers). Dotted lines: single-layer (one layer at a time, plotted by layer index). Diamond markers indicate the fewest layers from the back achieving $\geq$95\% of the full effect.}
    \label{fig:prefix_vs_suffix}
\end{figure}

To localize the CN effect across the network, we compare three layerwise interventions (\Cref{fig:prefix_vs_suffix}): (i)~\emph{suffix}, applying CN weights to only the last $k$ layers; (ii)~\emph{prefix}, applying CN weights to only the first $k$ layers; and (iii)~\emph{single-layer}, applying CN weights to one layer at a time. For each condition, we generate $N=120$ valid DAT samples at $T=1.0$ and report the percentage of the full CN effect recovered in terms of DAT scores.

\paragraph{Suffix.} Applying CN weights to a suffix of layers (layers $k$ through $L{-}1$) recovers the full effect with roughly half the network. On average, 51\% of layers (from the back) are needed to reach ~100\% recovery, ranging from 29\% for Qwen-7B (last 8/28) to 81\% for Llama-8B (last 26/32). The remaining models fall in between: Llama-3B (last 14/28, 50\%), Phi-14B (last 20/40, 50\%), Qwen-14B (last 22/48, 46\%), and Phi-4B (last 17/32, 53\%). However, this analysis is only conducted on DAT scores, and it is unclear whether the last 51\% of layers are sufficient to recover 100\% of the scores on the AUT and Task Task.

\paragraph{Prefix.} Applying CN weights from the front (layers $0$ through $k$) is far less efficient. On average, 78\% of all layers must be included before the prefix condition reaches 95\% recovery. For Qwen-14B (48 layers), the prefix condition never reaches 95\% even when all layers are included (94\% at $k=48$). This asymmetry provides evidence that the CN effect is concentrated in later layers of the residual stream.

\paragraph{Single-layer.} No single layer is sufficient to recover the full CN effect. The best individual layers recover 50--72\% of the effect (e.g., Q7B layer~19: 72\%, L3B layer~15: 64\%, P14B layer~20: 60\%), with top contributors generally appearing in middle-to-late layers. The gap between the best single layer and the suffix provides evidence that the CN effect requires cooperation across multiple late layers, consistent with findings that representations of human-interpretable concepts or behaviors may span multiple layers \citep{sparse_cross_coders, open_problems_in_mech_interp}.

\section{Hyperparameter Sweeps} \label{sec:hyperparam_sweeps}

CreativityNeuro introduces two hyperparameters: the \emph{importance threshold} $\rho \in (0, 1]$, which controls the fraction of parameters selected by the importance mask, and the \emph{scaling factor} $\alpha > 0$, which controls the magnitude of the weight perturbation applied to masked parameters. To identify effective $(\rho, \alpha)$ configurations for each model, we conduct systematic grid sweeps evaluated on the DAT. For each model, we generate CN weight masks using six different prompt sets: \textsc{dat}, \textsc{story}, \textsc{ideation}, \textsc{problem}, \textsc{openended}, and \textsc{minimal}. Each prompt set produces a separate importance mask. We then evaluate every combination of keep ratio $\rho \in \{0.01, 0.05, 0.1, 0.2\}$ and scaling factor $\alpha \in \{0.1, 0.5, 1.0, 2.0\}$ at three sampling temperatures $T \in \{0.9, 1.0, 1.2\}$, with $\text{top-}k = 0$ and $\text{top-}p = 1.0$ (i.e., untruncated sampling). For LLaMA-3.1-8B and Phi-3.5-mini, after initial experiments revealed that $\alpha > 0.5$ were too high, we additionally tested a finer alpha grid $\alpha \in \{0.01, 0.05, 0.075, 0.1, 0.5, 1.0\}$ to probe the conservative regime identified by \citet{Christ2025MathPasses}. Each configuration generates $n = 120$ valid DAT samples (10-word lists with all words present in the GloVe vocabulary).

\begin{figure}[p]
    \centering
    \includegraphics[width=\textwidth]{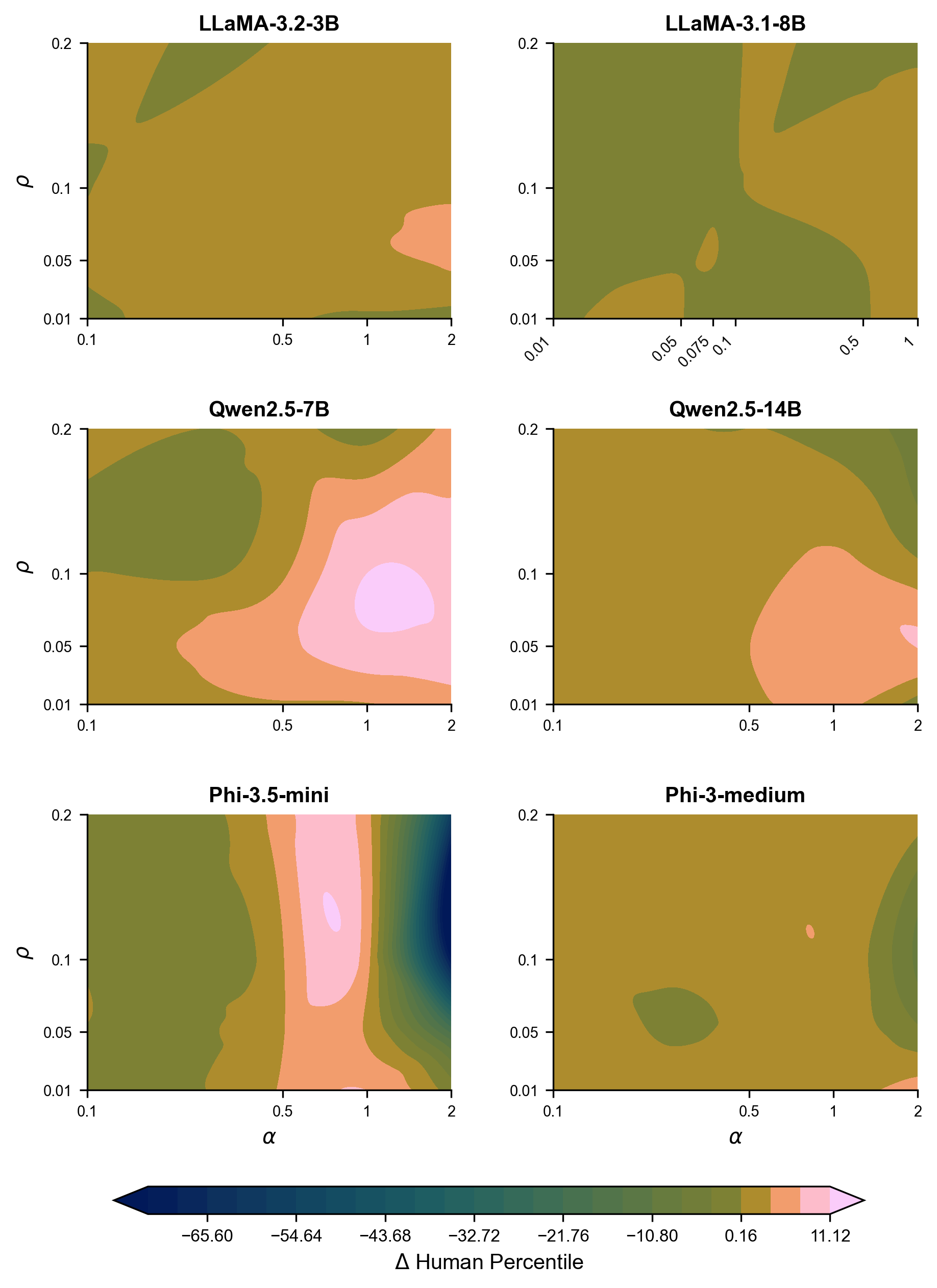}
    \caption{\textbf{Sensitivity to $(\alpha, \rho)$ --- DAT prompt set.} $\Delta$\,Percentile averaged across three temperatures ($T \in \{0.9, 1.0, 1.2\}$) for each model. Color scale is shared across panels and centered at zero.}
    \label{fig:sensitivity_dat}
\end{figure}

\begin{figure}[p]
    \centering
    \includegraphics[width=\textwidth]{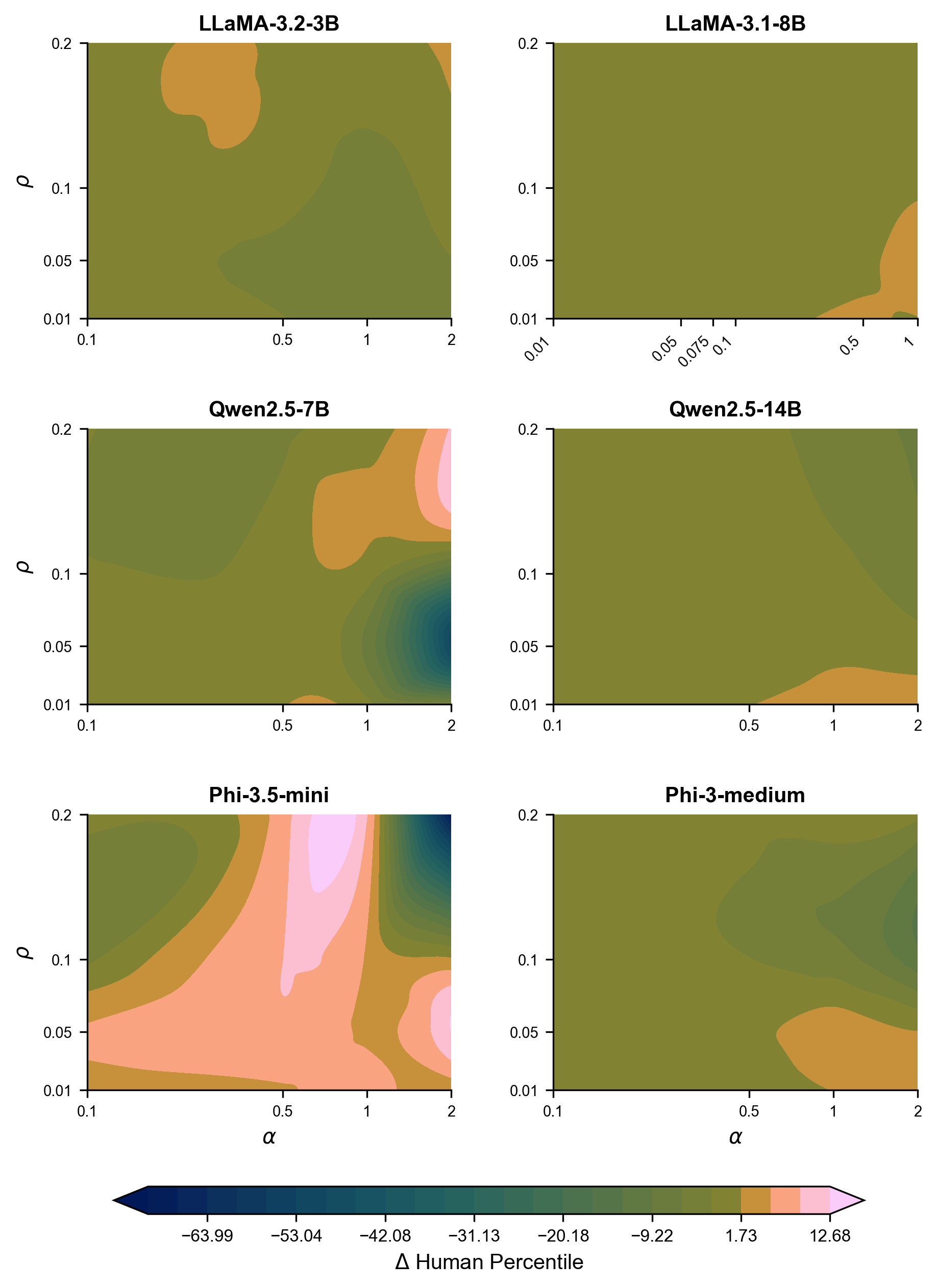}
    \caption{\textbf{Sensitivity to $(\alpha, \rho)$ --- Story prompt set.} Mean $\Delta$DAT averaged across three temperatures ($T \in \{0.9, 1.0, 1.2\}$) for each model. Color scale is shared across panels and centered at zero.}
    \label{fig:sensitivity_story}
\end{figure}

\begin{figure}[p]
    \centering
    \includegraphics[width=\textwidth]{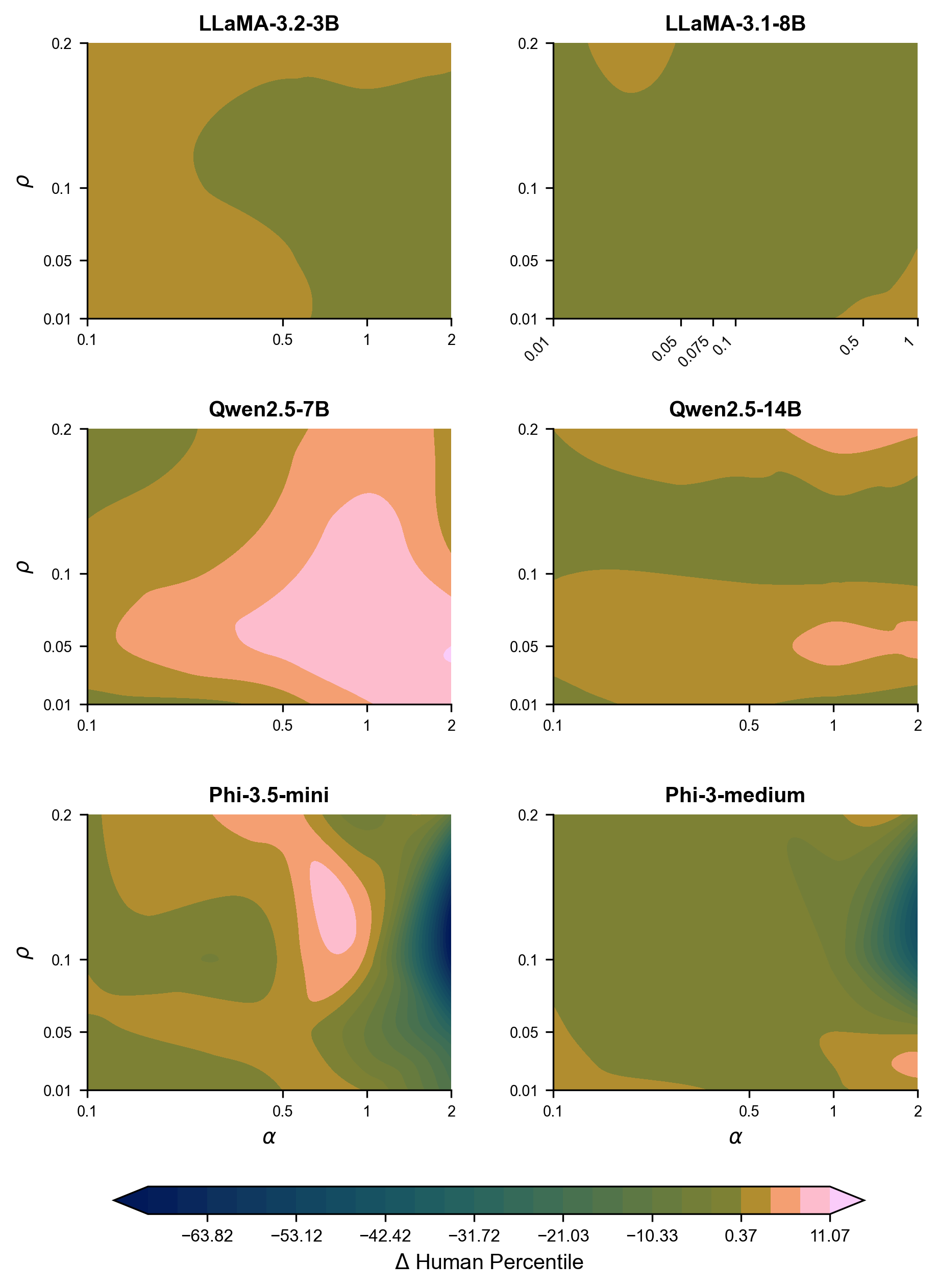}
    \caption{\textbf{Sensitivity to $(\alpha, \rho)$ --- Ideation prompt set.} Mean $\Delta$DAT averaged across three temperatures ($T \in \{0.9, 1.0, 1.2\}$) for each model. Color scale is shared across panels and centered at zero.}
    \label{fig:sensitivity_ideation}
\end{figure}

\begin{figure}[p]
    \centering
    \includegraphics[width=\textwidth]{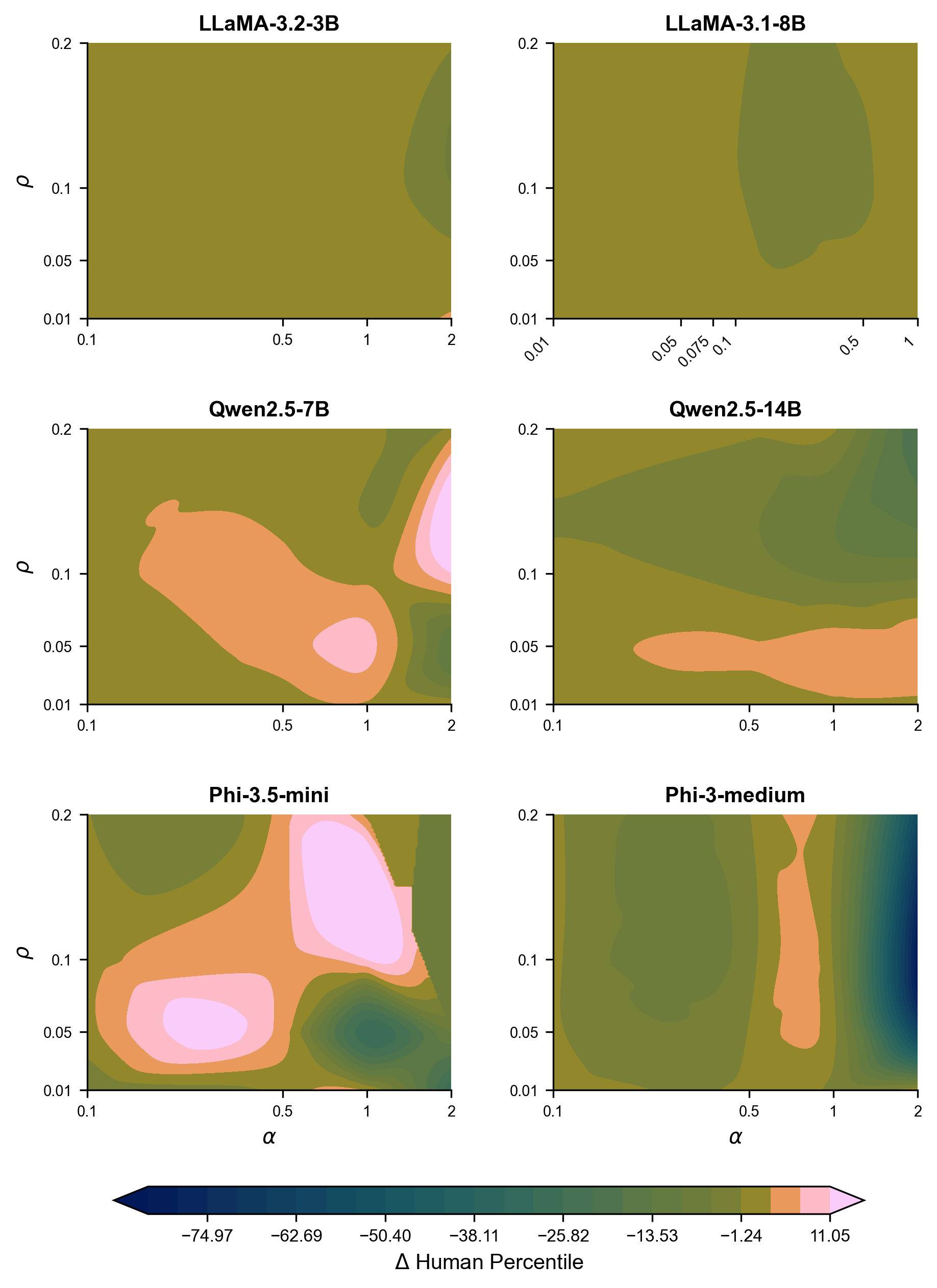}
    \caption{\textbf{Sensitivity to $(\alpha, \rho)$ --- Problem prompt set.} Mean $\Delta$DAT averaged across three temperatures ($T \in \{0.9, 1.0, 1.2\}$) for each model. Color scale is shared across panels and centered at zero.}
    \label{fig:sensitivity_problem}
\end{figure}

\begin{figure}[p]
    \centering
    \includegraphics[width=\textwidth]{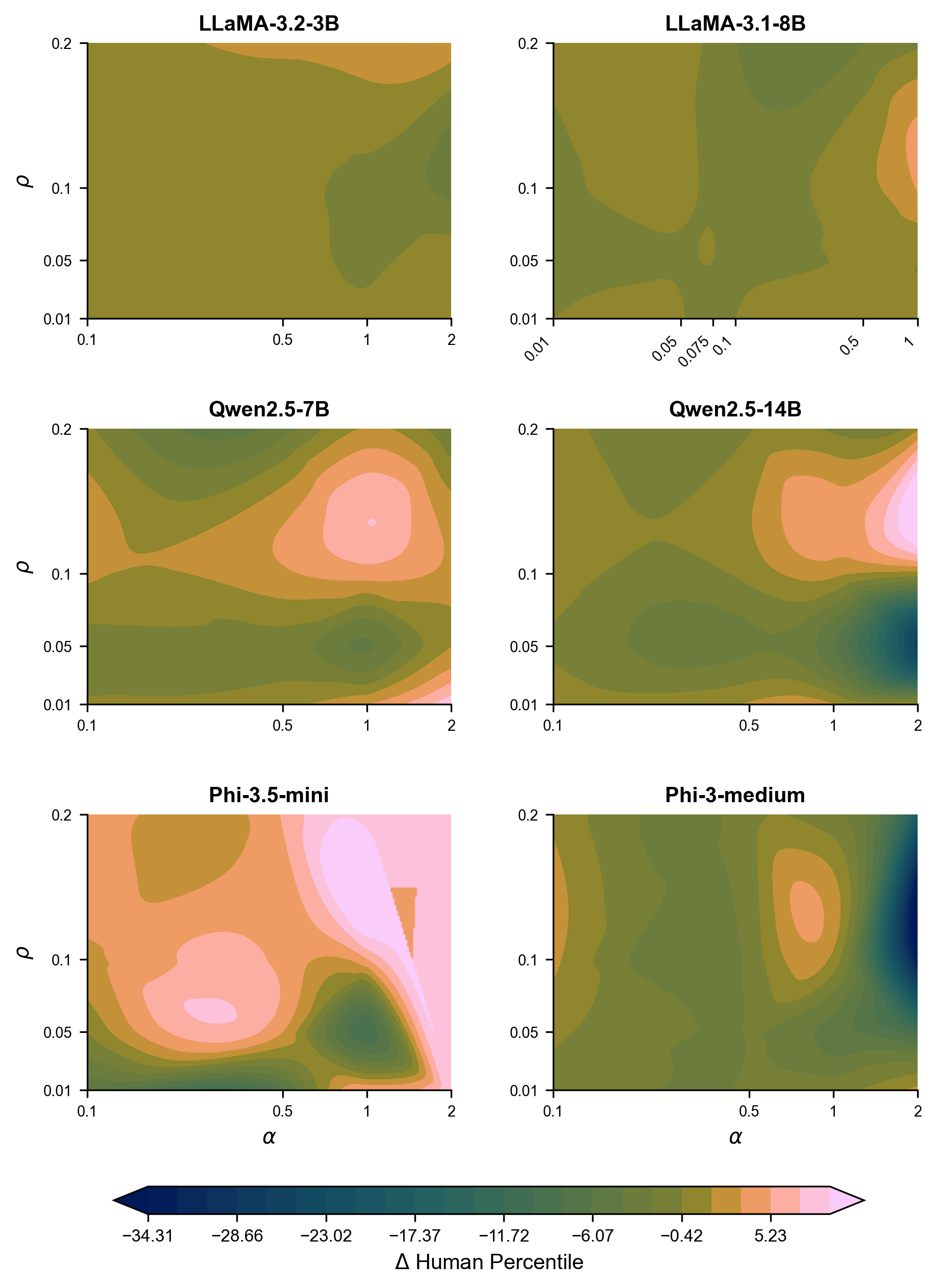}
    \caption{\textbf{Sensitivity to $(\alpha, \rho)$ --- Open-ended prompt set.} Mean $\Delta$DAT averaged across three temperatures ($T \in \{0.9, 1.0, 1.2\}$) for each model. Color scale is shared across panels and centered at zero.}
    \label{fig:sensitivity_openended}
\end{figure}

\begin{figure}[p]
    \centering
    \includegraphics[width=\textwidth]{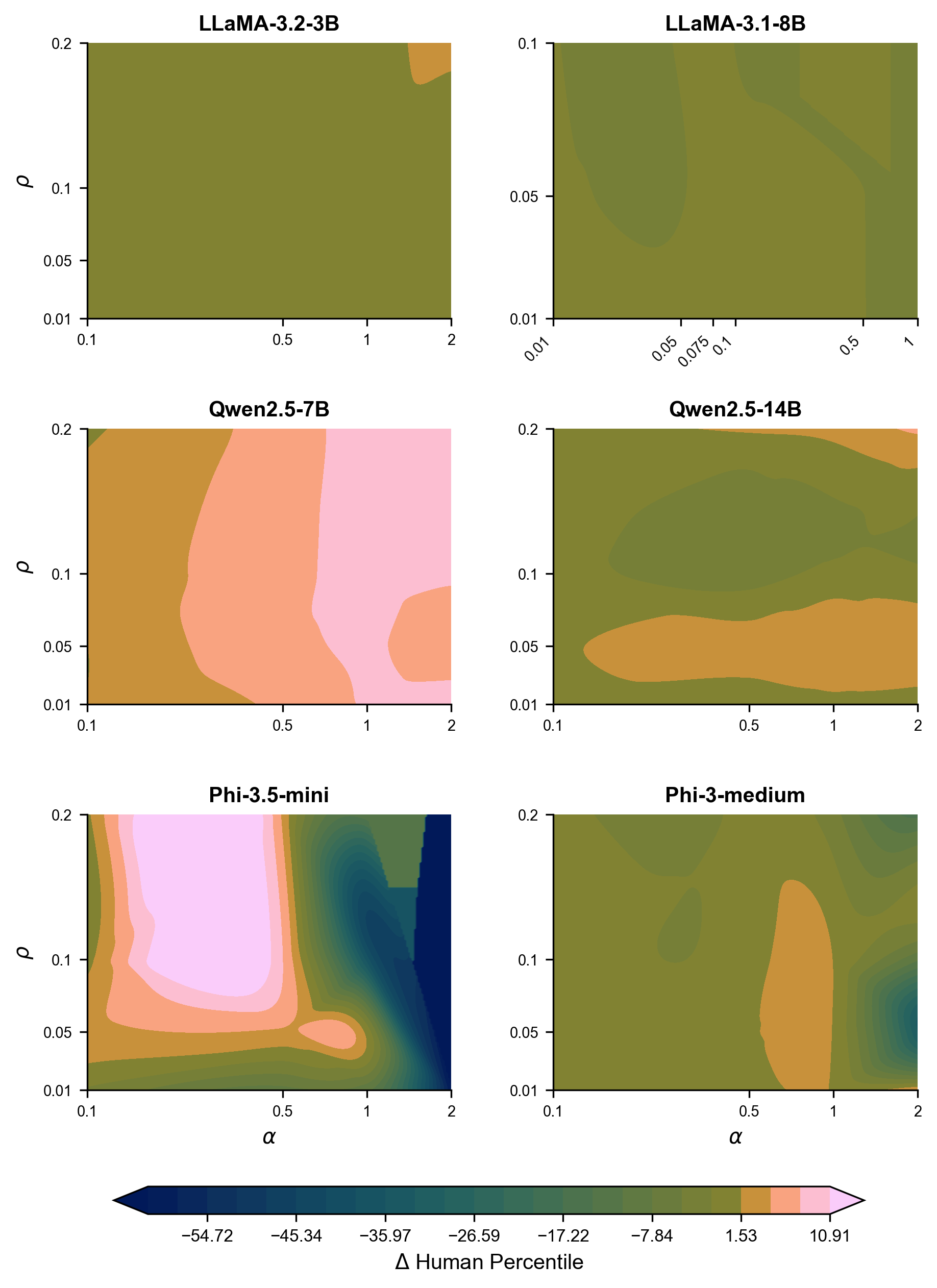}
    \caption{\textbf{Sensitivity to $(\alpha, \rho)$ --- Minimal prompt set.} Mean $\Delta$DAT averaged across three temperatures ($T \in \{0.9, 1.0, 1.2\}$) for each model. Color scale is shared across panels and centered at zero.}
    \label{fig:sensitivity_minimal}
\end{figure}

For each $(\rho, \alpha, T)$ triple, we compute the mean DAT scores for the CreativityNeuro ($\overline{\text{DAT}}_{\text{CN}}$) and baseline ($\overline{\text{DAT}}_{\text{base}}$) models, then convert each to a human percentile using the distribution from \citet{Wang2025LargeIdeas}. \Cref{fig:sensitivity_dat,fig:sensitivity_story,fig:sensitivity_ideation,fig:sensitivity_problem,fig:sensitivity_openended,fig:sensitivity_minimal} show $\Delta$\,Percentile $= P(\overline{\text{DAT}}_{\text{CN}}) - P(\overline{\text{DAT}}_{\text{base}})$ (averaged across temperatures) as a function of $(\alpha, \rho)$ for each of the six prompt sets, across all six models. Positive values indicate that CN moves the model's DAT performance upward in the human score distribution. 

\section{Evaluation Prompts} \label{sec:eval_prompts}
In this section, we share the full prompts used to generate and/or post-process the stimuli for the DAT, AUT, and TT.

\subsection{Divergent Association Task (DAT)}

The DAT prompt instructs models to generate 10 maximally dissimilar nouns:

\begin{promptbox}[DAT Prompt]
You will be asked to name exactly 10 English nouns.\newline
Output exactly 10 words, separated by commas, and nothing else.\newline
Return ONLY: word1, word2, ..., word10\newline
DO NOT write explanations. DO NOT think step by step.\newline\newline
Be as original and unusual as possible. Avoid common or closely related words.\newline
Now name 10 English nouns that are as different from each other as possible.
\end{promptbox}

\subsection{Alternative Uses Test (AUT)}

The AUT uses a system prompt and a user prompt, with \texttt{\{object\}} replaced by one of 3 standard objects: \emph{brick}, \emph{paperclip}\emph{fork}.

\begin{promptbox}[AUT System Prompt]
You are participating in a creativity test. Your task is to generate creative, unusual, and original uses for common objects. Be imaginative and think outside the box.
\end{promptbox}

\begin{promptbox}[AUT User Prompt]
List 5 creative and unusual alternative uses for a \{object\}.\newline
Be specific and creative. List each use on a new line, numbered 1 through 5.\newline
Only list the uses, no explanations.
\end{promptbox}

\subsection{Task Task (TT)}

The Task Task uses a system prompt and a user prompt with three in-context examples from \citet{chu2024task}.

\begin{promptbox}[TT System Prompt]
You are a creative game show designer. Your task is to invent fun, original, and entertaining challenges that would be exciting for contestants to attempt and audiences to watch.
\end{promptbox}

\begin{promptbox}[TT User Prompt]
You've been recruited to help design challenge tasks for a new game show! Your job is to come up with a new creative, silly, and fun task for humans to solve.\newline\newline
Here are a few example tasks:\newline\newline
1.\ Your goal is to: Throw a teabag into a mug from the farthest distance. You can use: Anything you can reasonably expect to find in a house, garage, and garden shed.\newline\newline
2.\ Someone has squeezed all of the toothpaste out of the toothpaste tube. Your goal is to: Get as much of the original toothpaste back into the empty toothpaste tube as possible. You can use: Anything you can reasonably expect to find in a bathroom.\newline\newline
3.\ Your goal is to: Transfer water between two fishbowls using only the supplied items. You cannot move the fishbowls. You can use: a chocolate bar, a rubber glove, a baguette, a snorkel, a cardboard tube, and a plate of pasta.\newline\newline
Now it's your turn! Create your own creative, silly, and fun task for future participants to solve. Specify the goal, scoring criteria, and any materials or constraints. Describe it in exactly 3--4 sentences as a short paragraph --- do not use lists or bullet points. Do not comment on how entertaining, creative, or fun the task would be.
\end{promptbox}

\subsection{Task Task Post-Processing} \label{sec:tt_postprocessing}

Raw Task Task generations are post-processed using GPT-4o to normalize formatting before human evaluation.

\begin{promptbox}[TT Post-Processing System Prompt]
You are an editor cleaning game show challenge descriptions for a human evaluation study.\newline\newline
Make these MINIMAL edits:\newline\newline
1. REMOVE any task title at the start (e.g., `In "Baking Bonanza,"'). After removing, capitalize the first word of the remaining text.\newline\newline
2. Convert ALL third-person references to SECOND PERSON (e.g., ``contestants must'' $\rightarrow$ ``you must'', ``the player'' $\rightarrow$ ``you'').\newline\newline
3. REMOVE any metacommentary sentences (e.g., ``This creative juggling act combines laughs with a dash of physics comedy.'').\newline\newline
CRITICAL RULES:\newline
- Do NOT summarize, condense, or shorten the task description\newline
- Do NOT change the creative content or game mechanics\newline
- Do NOT add any text --- only edit or remove\newline
- Output ONLY the cleaned description with no preamble or explanation
\end{promptbox}

\begin{promptbox}[TT Post-Processing User Prompt]
Clean this game show challenge description: \{text\}
\end{promptbox}

\section{Novelty--Utility Tradeoff Analysis} \label{sec:novelty_utility}

Among baseline AUT stimuli, originality and utility are negatively correlated ($r = -.49$, $p < 10^{-7}$), as are surprise and utility ($r = -.64$, $p < 10^{-13}$). As predicted by fundamental novelty--utility tradeoffs established in the creativity literature \citep{Varshney2019MathematicalCreativity}, increases in subjective ratings of novelty tend to be accompanied by decreases in perceived utility. Therefore, CN's utility reduction is an expected byproduct of steering towards high-novelty responses, rather than an independent failure mode. To confirm this, we perform a hypervolume (HV) analysis \citep{hypervolume} over all responses in the three-dimensional rating space (originality, surprise, utility) and find that the CN HV indicator is 15.2\% larger than baseline ($p = .18$), suggesting CN obtains a moderate (but non-significant) gain in Pareto efficiency as well.

\section{Disclosure of Large Language Model Usage}
In this paper, large language models (LLMs) were used to assist in the code implementation, plotting and figure generation, reporting (but not analysis) of experiment results, and for comprehensive surveys of related work. 